\documentclass{article}

\usepackage[final]{nips_2016}

\usepackage[utf8]{inputenc}
\usepackage[T1]{fontenc}
\usepackage{hyperref}
\usepackage{url}
\usepackage{booktabs}
\usepackage{amsfonts}
\usepackage{nicefrac}
\usepackage{microtype}
\usepackage{authblk}
\usepackage{color}
\usepackage{verbatim}
\usepackage{graphicx}
\usepackage{caption}
\usepackage{subcaption}
\usepackage{amsmath}

\title{Robust Variational Inference}
\newcommand{\ELBO}{\mathcal L}

\newcommand{\MEAN}{\mathbb E}
\newcommand{\MEANq}{\mathbb E_{q(z_i | x_i, \phi)}}

\begin{document}

\author[1]{\textbf{Michael Figurnov}}
\author[1]{\textbf{Kirill Struminsky}}
\author[1,2]{\textbf{Dmitry Vetrov}}

\affil[1]{National Research University Higher School of Economics}
\affil[2]{Yandex}
\affil[ ]{\tt\small michael@figurnov.ru \ k.struminsky@gmail.com \ vetrovd@yandex.ru}

\maketitle
\begin{abstract}
Variational inference is a powerful tool for approximate inference. However, it mainly focuses on the evidence lower bound as variational objective and the development of other measures for variational inference is a promising area of research. This paper proposes a robust modification of evidence and a lower bound for the evidence, which is applicable when the majority of the training set samples are random noise objects. We provide experiments for variational autoencoders to show advantage of the objective over the evidence lower bound on synthetic datasets obtained by adding uninformative noise objects to MNIST and OMNIGLOT. Additionally, for the original MNIST and OMNIGLOT datasets we observe a small improvement over the non-robust evidence lower bound.
\end{abstract}

\section{Introduction}
One of the common approaches to approximate Bayesian inference, especially popular for large-scale models, is maximization of variational lower bound on the model's evidence. In recent years several new objectives were proposed. \citet{burda2016} gave a tighter evidence lower bound (ELBO) which exploits multiple samples from the approximating distribution. \citet{li2016} extended traditional variational inference with Renyi divergence-inspired evidence bounds. These bounds were shown to be more effective for certain unsupervised learning problems.

In this paper we consider the problem of unsupervised learning for data which is known to contain many uninformative (noise) objects. First, we replace the traditional log-evidence $\sum_i \log p(x_i)$ with a robust counterpart $\sum_i \log (\varepsilon + p(x_i))$.
This function ignores the objects with low evidence $p(x_i) \ll \varepsilon$. Next, we derive a variational lower bound on robust model evidence which also shares the same robustness property. We show that by maximizing this lower bound we can successfully train variational autoencoders even in the scenarios where the noise objects comprise the majority of the training dataset.
An alternative approach, proposed by \citet{wang2016reweighted}, is to reweight the per-object likelihood terms with additional local latent variables.
In future, we plan to compare to this approach.

\section{Robust variational inference}

In what follows we consider a parametric latent variable model $p(x, z | \theta) = p(x | z, \theta) p(z) $ with local latent variables $z$ and parameter $\theta$. We derive a lower bound for the robust evidence and study its properties. Finally, following \citep{kingma2014}, we propose a training procedure for variational autoencoders with the robust evidence lower bound.

\subsection{Robust evidence lower bound}

We start off with the robust log-evidence $\sum_{i = 1}^N \log (\varepsilon + p(x_i | \theta))$.
First, we rewrite it for each sample:

\begin{equation}
\log \left[ \varepsilon + p(x_i | \theta) \right] = \log \left[ \MEANq \left( \varepsilon + \frac{p(x_i, z_i | \theta)}{q(z_i | x_i, \phi)} \right) \right]
\end{equation}

and then apply Jensen's inequality to obtain the robust evidence lower bound $\ELBO_{\varepsilon}(X, \theta, \phi)$

\begin{equation}
\sum_{i = 1}^N \log (\varepsilon + p(x_i | \theta)) \geq \sum_{i = 1}^{N} \MEAN_{q(z_i | x_i, \phi)} \log \left[ \varepsilon + \frac{p(x_i, z_i | \theta)}{q(z_i | x_i, \phi )} \right] = \ELBO_{\varepsilon}(X, \theta, \phi).
\end{equation}

The robust evidence bound is tight when the variational distribution $q(z_i | x_i, \phi)$ is the true posterior $p(z_i | x_i, \theta)$.

This objective exhibits robustness to the objects with the low value of $\frac{p(x_i, z_i | \theta)}{q(z_i | x_i, \phi )}$.
For a fixed sample $(x_i, z_i)$ explicit computation gives

\begin{equation}
\nabla \log\left[ \varepsilon + \frac{p(x_i, z_i| \theta)}{q(z_i | x_i, \phi)} \right] =
\left(
\frac{p(x_i, z_i| \theta)}{q(z_i | x_i, \phi)} \left[\varepsilon + \frac{p(x_i, z_i| \theta)}{q(z_i | x_i, \phi)} \right]^{-1}
\right)\nabla \log \frac{p(x_i, z_i | \theta)}{q(z_i | x_i, \phi)}.
\end{equation}

Therefore, the stochastic gradient of the robust evidence lower bound has the same direction as the gradient of the non-regularized ELBO $\mathcal{L}$ for this sample $(x_i, z_i)$:

\begin{equation}
    \nabla \log \frac{p(x_i, z_i | \theta)}{q(z_i | x_i, \phi )}.
\end{equation}

When $\frac{p(x_i, z_i | \theta)}{q(z_i | x_i, \phi )} \ll \varepsilon$, the scalar factor before this gradient is close to zero and the sample does not contribute to the parameter update.
On the other hand, when $\frac{p(x_i, z_i | \theta)}{q(z_i | x_i, \phi )} > \varepsilon$, the factor lies in $[ \frac{1}{2}, 1)$ and we obtain almost the same update as for the non-regularized ELBO.

To benefit from the robustness one has to choose $\varepsilon$ carefully. Underestimating $\varepsilon$ results in poor regularization, overestimating $\varepsilon$ results in significant distortion of the evidence.
It is natural choose $\varepsilon$ value to be comparable to the log-likelihood of the dataset.
We propose to use a \textit{dynamically changing} value for $\varepsilon$, specifically a multiple of the mean evidence lower bound:

\begin{equation}
\varepsilon = \alpha \exp{ \left( \frac{\ELBO(X, \theta, \phi)}{|X|} \right) }.
\label{eqn:eps}
\end{equation}

Here $\alpha > 0$ controls the regularization effect.
As $\alpha \rightarrow 0$ the robust evidence lower bound converges to ELBO.

\subsection{Training procedure}

To train the model, a stochastic gradient based optimizer is used to maximize the objective function. We use Gaussian latent variables and employ reparametrization trick to obtain the gradient estimates. Firstly, we train the model for one epoch with the evidence lower bound as an objective and initialize $\log \varepsilon$ as the mean ELBO value at the first epoch. Secondly, we fix $\log \alpha$ and then train the model using the robust evidence lower bound as an objective. After each gradient step we update $\log \varepsilon$ with the mean ELBO of the previous batch using exponential smoothing. Moreover, after each epoch we update $\log \varepsilon$ with the mean ELBO from the previous epoch.

\section{Experiments}

In our experiments we used a model with the following architecture: fully-connected encoder and decoder had two hidden layers with 200 units, stochastic layer had 50 hidden units. Parametric ReLU \citep{he2015delving} activation units were used for the deterministic layers.

We used Adam (\cite{kingma2015}) with parameters $\beta_1 = 0.99, \beta_2 = 0.999, \epsilon = 10^{-4}$ for objective maximization. Each model was trained for 1000 epochs with a fixed learning rate of $10^{-3}$. Batch size was set to 200. The following rule was used to update $\varepsilon$ after processing of each batch: $\log \varepsilon_{new} = 0.99 \log \varepsilon_{old} + 0.01 \log \varepsilon $, where $\varepsilon$ is estimated using eqn. \eqref{eqn:eps}.

In the first experiment we compared the robust variational autoencoders with autoencoders on two synthetic datasets. We used MNIST and OMNIGLOT \citep{lake2015human} as real-world base sets, and then added uninformative data points, i.e. $28*28$ images with each pixel's intensity equal to the mean pixel intensity of the original dataset. Due to dynamic binarization \citep{burda2016} these data points act as noise. We varied the relation of the number of the original data points to the number of noise data points from 2:1 to 1:2.

To evaluate the models performance we computed mean log-likelihood estimate over 200 samples on MNIST and OMNIGLOT test sets (without any noise).
The range of $\log \alpha$ was selected empirically: we started with $\log \alpha = -50$ and then increased it to find the optimal value with respect to the test likelihood. 

\begin{figure}
\centering
\includegraphics[width=0.9\linewidth]{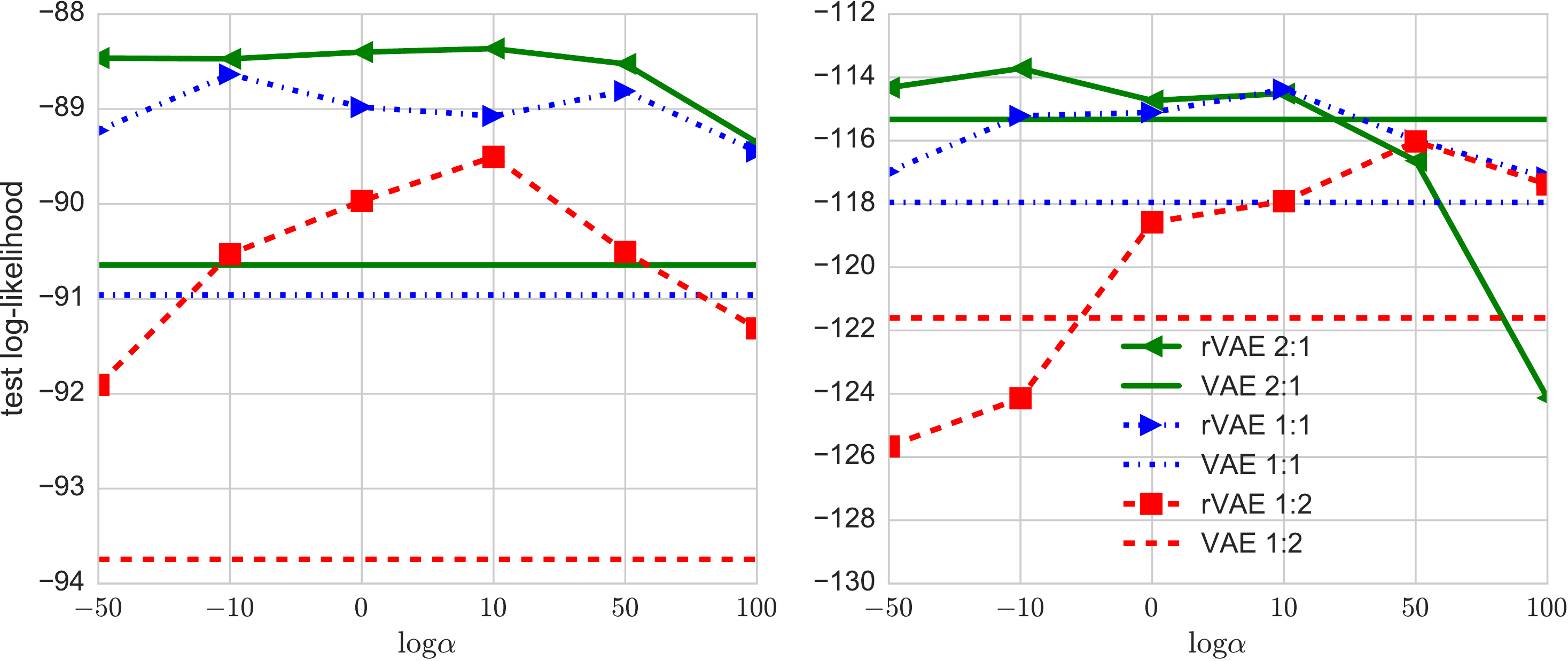}
\caption{\textbf{Left}: noisy MNIST, \textbf{right}: noisy OMNIGLOT. The proportion of (original:noise) data points is varied from 2:1 to 1:2. We compared test log-likelihood of the \textit{original dataset} for variational autoencoders (VAE) and the proposed robust autoencoders (rVAE) with different regularization parameters $\alpha$ (note that the x-axis is not uniform). rVAE successfully ignores the noise data points while VAE's quality degrades significantly.}
\label{fig:a}
\end{figure}

\begin{figure}
\centering
\begin{subfigure}[b]{0.4\linewidth}
\includegraphics[width=\linewidth]{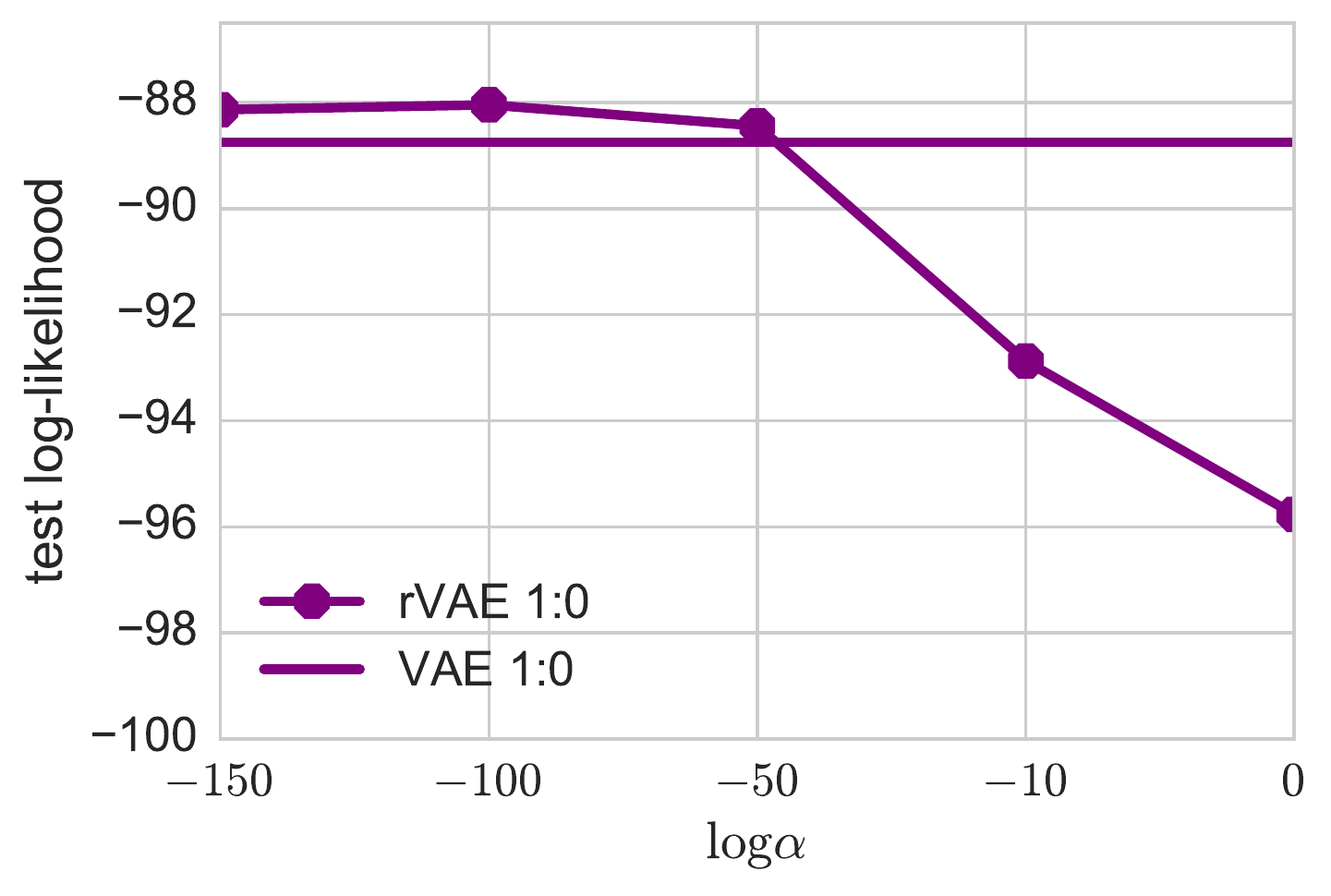}
\end{subfigure}
~
\begin{subfigure}[b]{0.4\linewidth}
\includegraphics[width=\linewidth]{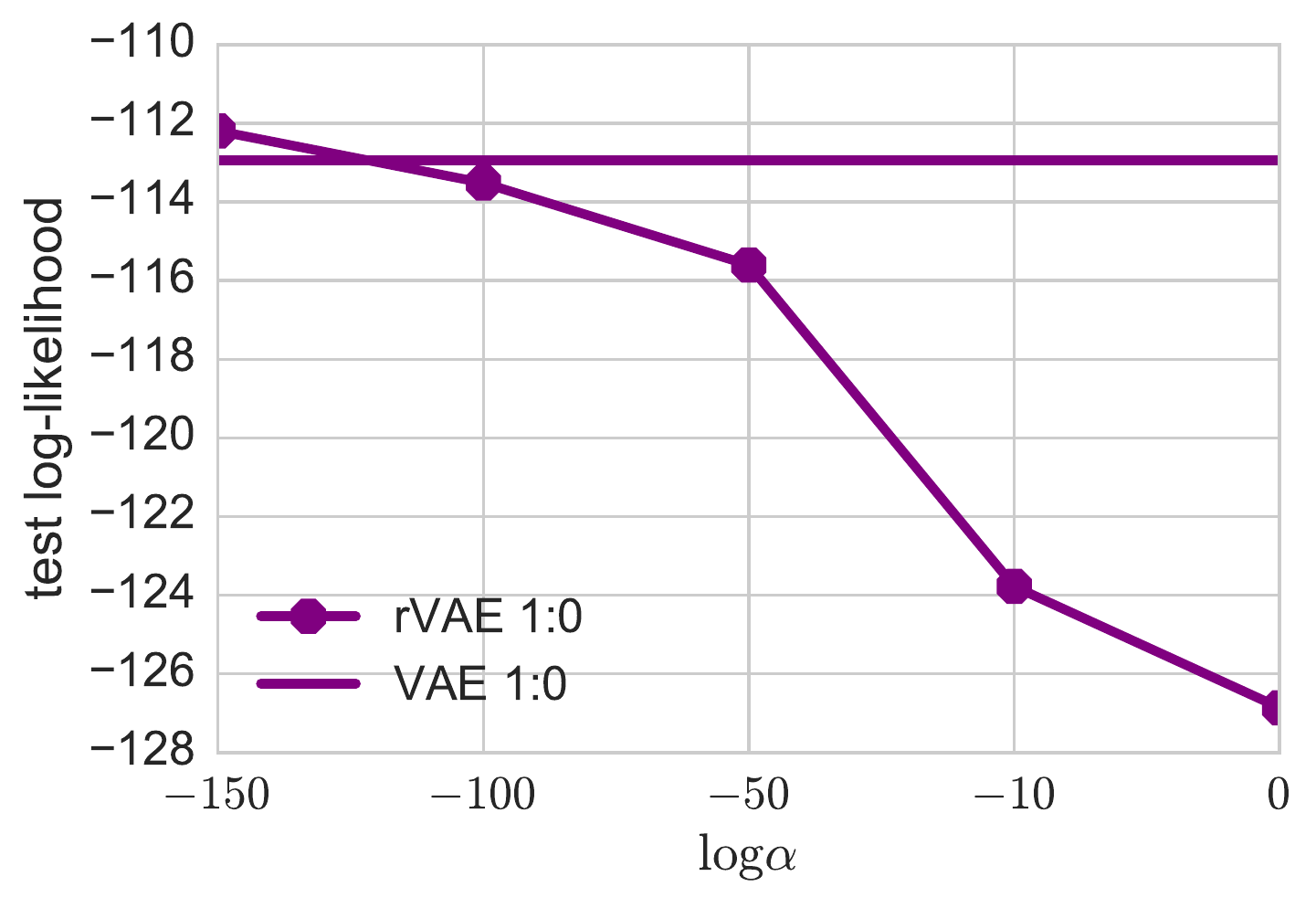}
\end{subfigure}
\caption{\textbf{Left}: MNIST, \textbf{right}: OMNIGLOT. Log-likelihood estimates for the robust autoencoder and the variational autoencoder trained without the synthetic noise. In this setting, choosing a very small value of $\log \alpha$ results in a regularization effect leading to a small improvement over VAE ($0.7$ nats for MNIST, $0.74$ nats for OMNIGLOT).
%$0.74$ for $\log \alpha = -150$ on OMNIGLOT,  $0.70$ and $0.61$ for $\log \alpha = -100$ and $\log \alpha = -150$ on MNIST correspondingly).
}
\label{fig:b}
\end{figure}

Results of the first experiment are shown in Figure~\ref{fig:a}. The robust autoencoder managed to fit the data despite noise. At the same time, VAEs test log-likelihood significantly decreased as the fraction of noise increased. The optimal value of $\alpha$ depends on base dataset and fraction of noise. For example, for OMNIGLOT the best $\log \alpha$ increased monotonically  with the fraction of noise. However, for MNIST there was no such simple pattern. 

In the second experiment we have compared rVAEs and VAEs on datasets without synthetic noise. We used the same network architecture and optimization approach. Test log-likelihoods for MNIST and OMNIGLOT datasets are presented in Figure~\ref{fig:b}. In this setting the best results were achieved when $\alpha$ almost coincided with zero. We observed a small improvement of the robust autoencoder over the variational autoencoder, suggesting that robust VAE provides a beneficial regularization effect.

\section{Conclusion}
We presented a new variational objective for approximate inference and showed its advantage in the training setting where noisy objects comprise the majority of a dataset. Additionally, the proposed robust variational objective provides small regularization effect on datasets without any artificial noise. In future we plan to incorporate the regularization parameter into the probabilistic model, design a procedure for automatic selection of the parameter and evaluate the model on a real-world noisy dataset.

\textbf{Acknowledgments.}
This work was supported by RFBR project No. 15-31-20596 (mol-a-ved) and by Microsoft: Moscow State University Joint Research Center (RPD 1053945).

{\small
\bibliographystyle{IEEEtranSN}
\bibliography{main}
}
\end{document}